\documentclass[conference]{IEEEtran}
\IEEEoverridecommandlockouts

\usepackage{cite}
\usepackage{amsmath,amssymb,amsfonts}
\usepackage{algorithmic}
\usepackage{graphicx}
\usepackage{textcomp}
\usepackage{xcolor}
\usepackage{array}
\usepackage{hyperref}
\usepackage{booktabs}
\def\BibTeX{{\rm B\kern-.05em{\sc i\kern-.025em b}\kern-.08em
    T\kern-.1667em\lower.7ex\hbox{E}\kern-.125emX}}
\begin{document}
\title{An Automated Machine Learning Framework for Surgical Suturing Action Detection under Class Imbalance\\

\thanks{This research was fully funded by EPSRC, UK. With the Grant Reference EP/Y017307/1.}
\thanks{* Corresponding Author}
}

\author{\IEEEauthorblockN{1\textsuperscript{st} Baobing Zhang*}
\IEEEauthorblockA{\textit{School of Engineering and Physical Sciences} \\
\textit{Heriot Watt University}\\
Edinburgh EH14 4AS, UK \\
B.Zhang@hw.ac.uk
}

\and
\IEEEauthorblockN{2\textsuperscript{st} Paul Sullivan}
\IEEEauthorblockA{\textit{School of Engineering and Physical Sciences} \\
	\textit{Heriot Watt University}\\
	Edinburgh EH14 4AS, UK \\
	P.Sulliva@hw.ac.uk
}

\and
\IEEEauthorblockN{3\textsuperscript{st} Benjie Tang}
\IEEEauthorblockA{\textit{Surgical Skills Centre, Dundee Institute for Healthcare
		Simulation} \\
\textit{Ninewells Hospital and Medical School, University
	of Dundee}\\
Dundee, UK \\
b.tang@dundee.ac.uk}

\and
\IEEEauthorblockN{4\textsuperscript{st} Ghulam Nabi}
\IEEEauthorblockA{\textit{Surgical Skills Centre, Dundee Institute for Healthcare
		Simulation} \\
	\textit{Ninewells Hospital and Medical School, University
		of Dundee}\\
	Dundee, UK \\
	g.nabi@dundee.ac.uk}
	
\and
\IEEEauthorblockN{5\textsuperscript{st} Mustafa Suphi Erden*}
\IEEEauthorblockA{\textit{School of Engineering and Physical Sciences} \\
	\textit{Heriot Watt University}\\
	Edinburgh EH14 4AS, UK \\
	@hw.ac.uk
}


}


\maketitle

\begin{abstract}
In laparoscopy surgical training and evaluation, real-time detection of surgical actions with interpretable outputs is crucial for automated and real-time instructional feedback and skill development. Such capability would enable development of machine guided training systems. This paper presents a rapid deployment approach utilizing automated machine learning methods, based on surgical action data collected from both experienced and trainee surgeons. The proposed approach effectively tackles the challenge of highly imbalanced class distributions, ensuring robust predictions across varying skill levels of surgeons. Additionally, our method partially incorporates model transparency, addressing the reliability requirements in medical applications. Compared to deep learning approaches, traditional machine learning models not only facilitate efficient rapid deployment but also offer significant advantages in interpretability. Through experiments, this study demonstrates the potential of this approach to provide quick, reliable and effective real-time detection in surgical training environments. 
\end{abstract}

\begin{IEEEkeywords}
Bayesian optimization, reliable, bayesian learning, probabilistic models, trustworthy.
\end{IEEEkeywords}

\section{INTRODUCTION}

Laparoscopic surgery, as a minimally invasive technique, is now considered a crucial choice in modern surgery due to its advantages, such as minimal trauma and faster recovery \cite{madhok2022safety, basunbul2022recent}. However, the unique perspective and operation mode of laparoscopic surgery place higher demands on surgeons' hand-eye coordination and spatial awareness, where the precision of the operation directly impacts surgical quality and patient recovery outcomes \cite{stulberg2020association, sanchez2020comparative}. To improve surgeons' laparoscopic skills, laparoscopic surgical training platforms have been developed. These platforms simulate real surgical environments, providing trainees with a safe space to practice \cite{hong2021simulation, van2020bimanual}. However, most existing platforms rely on fixed training steps and manual feedback, lacking intelligent real-time feedback mechanisms, and are unable to dynamically track trainees' movements. A surgical action recognition system, however, has the potential to monitor and analyze trainees' actions in real-time, providing immediate feedback during training, which can help trainees correct errors promptly and improve training efficiency and accuracy \cite{hashimoto2018artificial}. Such real-time feedback is crucial for provision of automated and real-time feedback, to prepare the next-generation laparoscopy training systems that eliminate the need of expert monitoring \cite{larsen2009effect}.

In practical applications, surgical action recognition systems face several major challenges. First, class imbalance is a common issue in medical data, especially in surgical data, where key actions often have fewer samples, while routine actions are abundant. This imbalance can lead to model bias toward majority classes, diminishing the recognition performance for minority classes \cite{krawczyk2016learning, he2009learning, johnson2019survey}. Second, model interpretability is essential in medical applications. Surgeons and trainers require not only a highly accurate model but also one with a transparent decision-making process to ensure the rationality and credibility of the model's outputs \cite{rudin2019stop, tjoa2020survey}. However, current deep learning \cite{lecun2015deep} models often have complex decision processes and lack interpretability and reliability, while statistical model-based decision making process which provides thorough interpretability and formal reliability guarantee \cite{zhang2024bayesian}. In addition, model stability and robustness are critical for the successful application of a surgical action recognition system. With the presence of both complexity and diversity in medical data, models may exhibit inconsistent performance across different data and environments, making it essential for the system to maintain stability and reliability across varied scenarios \cite{esteva2021deep, topol2019high}. Lastly, real-time feedback is indispensable in surgical training, as an efficient surgical action recognition system must process actions in real-time to provide immediate feedback on the quality of trainees' movements \cite{aggarwal2010training, sinha2023current}.

This study aims to propose an efficient surgical action recognition system to address the aforementioned challenges, with particular focus on handling class imbalance, enhancing model interpretability and stability, and achieving real-time feedback. To this end, we incorporate automated machine learning technology, by using automated machine learning to automatically construct and optimize models, thereby avoiding the complex process of manual tuning, while enhancing model stability and generalization through ensemble learning methods. In model construction, we combine techniques such as sample resampling and weighted classification to address class imbalance, ensuring the significance of minority classes in model training. Moreover, to improve system transparency and interpretability, we utilize traditional machine learning algorithms and interpretability techniques, making the decision process understandable for both surgeons and trainees. Finally, we designed and implemented a real-time surgical action recognition system capable of timely action recognition during training, which can be used in an intelligent and dynamic evaluation tool for laparoscopic surgical training platforms. 

This paper is structured in the following manner: Section 2 reviews prior research, focusing on advancements in surgical action recognition, class imbalance handling, interpretability, and real-time feedback. Section 3 details the methodology for laparoscopic surgical action recognition, including class imbalance handling, hyperparameter optimization, and ensemble learning. Section 4 provides an overview of the experimental configuration and findings, analyzing the system's accuracy, stability, and feedback efficiency on surgical training data. Section 5 concludes with a summary of the main contributions and discusses the potential applications and future directions of this system in surgical training.

\section{Related Work}

Laparoscopy surgical action recognition is a crucial research area in surgical training, for provision of real-time feedback to trainees during procedures. Lea et al \cite{lea2016segmental}. proposed Segmental Spatiotemporal Convolutional Neural Networks (CNNs) for capturing fine-grained temporal information to perform action segmentation, achieving significant results in surgical action recognition. Similarly, Twinanda et al \cite{twinanda2016endonet}. developed EndoNet, an advanced neural network architecture specifically designed for laparoscopic surgical videos, capable of performing multitask recognition, including action classification. The multitask structure of EndoNet effectively improves the recognition of complex actions, offering a novel approach for action detection in laparoscopic videos. In surgical data processing, class imbalance is a common phenomenon. Chawla et al \cite{chawla2002smote}. proposed SMOTE (Synthetic Minority Over-sampling Technique), which generates synthetic minority samples to balance the dataset distribution and improve recognition of minority class actions. Additionally, the survey by He and Garcia \cite{he2009learning} provides a comprehensive review of various techniques for handling class imbalance, including oversampling, undersampling, and weighted methods, which play a critical role in enhancing minority class recognition.

In medical applications, model interpretability is essential for machine learning model adoption. Rudin \cite{rudin2019stop} emphasized the necessity of using interpretable models in high-stakes domains and recommended prioritizing interpretable models over black-box models to ensure higher transparency for clinicians and trainers. Moreover, the work by Doshi-Velez and Kim \cite{doshi2017towards} explores various techniques for enhancing model interpretability, presenting a framework for applying interpretable models in medical applications, providing theoretical support for the use of interpretable models in clinical scenarios. Regarding model stability, Breiman's random forest algorithm \cite{breiman2001random}, as an ensemble learning method, enhances model stability and generalization by combining multiple decision trees, making it widely applicable to medical data processing where high stability is required. To further improve model stability and automate the optimization process, Feurer et al. \cite{feurer2015efficient} developed an automated machine learning system that combines automated model selection and hyperparameter tuning, effectively reducing the complexity of manual tuning and improving model adaptability and stability on surgical data. In surgical training, real-time feedback systems assist trainees by providing immediate feedback during procedures. Salvador et al \cite{salvador2024effects} investigated the application of real-time visual feedback in laparoscopic training, examining its impact on novices' learning curves. The results showed that trainees receiving real-time feedback demonstrated higher precision in tissue handling skills, significantly shortening their learning curves. The introduction of real-time feedback enabled trainees to master essential skills more quickly, improving training efficiency.

In summary, current laparoscopy surgical action recognition systems, class imbalance handling techniques, model interpretability methods, stability-enhancing techniques, and real-time feedback systems have shown progress in surgical training, yet limitations remain. Addressing these gaps, this study aims to optimize surgical action recognition systems using automated machine learning, combining multiple techniques to improve model performance in accuracy, interpretability, stability, and real-time functionality, providing a more intelligent and efficient feedback mechanism for surgical training.

\begin{figure*}[h!]
	\centering
	\includegraphics[width=\textwidth]{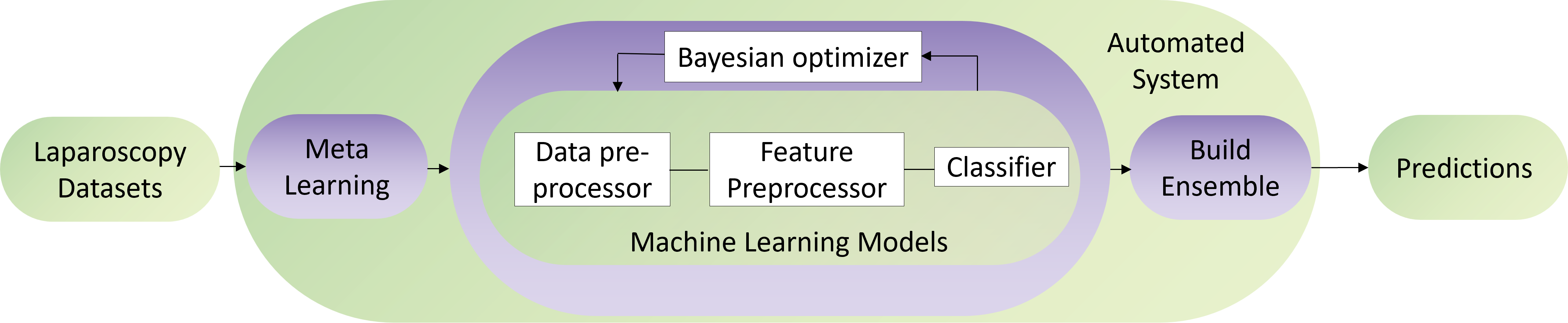}
	\caption{Overall AutoML workflow including meta learning warmstart for bayesian optimization efficient model selection and ensemble building for laparoscopy surgical suturing action detection. 
	}
	\label{framework}
\end{figure*}

\section{Method}

The focus of this study is on surgical action recognition within the context of laparoscopy training and evaluation. This task aims to generate accurate predictions for surgical actions from trajectory data in real-time, leveraging automated machine learning techniques to minimize the need for human intervention. In practical applications, the computational budget is defined by the specific requirements of laparoscopy training systems, including constraints on CPU time, memory usage, and latency, ensuring compatibility with the real-time feedback needs of surgical training platforms. These constraints are critical for deploying models in training environments where quick and reliable action recognition is essential for effective evaluation and feedback. Specifically, the automated machine learning for surgical action recognition can be formulated in the following manner:

Let \( i = 1, \ldots, n + m \), where \( x_i \in \mathbb{R}^d \) denotes a feature vector and \( y_i \in Y \) represents a paired response label. Given a training dataset \( D_{\text{train}} = \{(x_1, y_1), \ldots, (x_n, y_n)\} \) and a test dataset \( D_{\text{test}} = \{(x_{n+1}, y_{n+1}), \ldots, (x_{n+m}, y_{n+m})\} \), where samples are obtained from a consistent data structure, the goal is to derive automated predictions \( \hat{y}_{n+1}, \ldots, \hat{y}_{n+m} \) for the test set, given a loss function \( \mathcal{L}(\cdot, \cdot) \) and a resource budget. The loss of the solution is defined as \( \frac{1}{m} \sum_{j=1}^{m} \mathcal{L}(\hat{y}_{n+j}, y_{n+j}) \).


The formal definition of automated machine learning (AutoML) as a Combined Algorithm Selection and Hyperparameter optimization (CASH) problem was introduced by the AutoML approach in the AUTO-WEKA \cite{hall2009weka} system. In AutoML, a single machine learning algorithm may not always perform optimally across different datasets, making algorithm selection and hyperparameter optimization essential components of an AutoML system. The CASH problem integrates model selection and hyperparameter optimization into a single unified optimization task that can be solved using Bayesian optimization, aiming to identify the optimal algorithm and hyperparameter settings that minimize the loss function on a given dataset.

For the definition of CASH, let a set of algorithms \( \mathcal{A} = \{ A^{(1)}, \dots, A^{(R)} \} \), where each algorithm \( A^{(j)} \) has its own hyperparameter space \( \Lambda^{(j)} \). Given a training dataset \( D_{\text{train}} = \{ (x_1, y_1), \dots, (x_n, y_n) \} \), we partition it into \( K \) cross-validation folds, consisting of validation sets \( \{ D_{\text{valid}}^{(1)}, \dots, D_{\text{valid}}^{(K)} \} \) and corresponding training subsets \( \{ D_{\text{train}}^{(1)}, \dots, D_{\text{train}}^{(K)} \} \), where \( D_{\text{train}}^{(i)} = D_{\text{train}} \setminus D_{\text{valid}}^{(i)} \) for \( i = 1, \dots, K \). The loss function is donated as \( \mathcal{L}(A_{\lambda}^{(j)}, D_{\text{train}}^{(i)}, D_{\text{valid}}^{(i)}) \), representing the loss of algorithm \( A^{(j)} \) with hyperparameters \( \lambda \) when evaluated on \( D_{\text{valid}}^{(i)} \) and trained on \( D_{\text{train}}^{(i)} \). In the context of the CASH framework, the goal is to determine the best combination of model selection and hyperparameter tuning, aiming to reduce the average validation error across all cross-validation folds, formulated as:
\[
A^*, \lambda^* = \arg\min_{A^{(j)} \in \mathcal{A}, \lambda \in \Lambda^{(j)}} \frac{1}{K} \sum_{i=1}^{K} \mathcal{L}(A_{\lambda}^{(j)}, D_{\text{train}}^{(i)}, D_{\text{valid}}^{(i)}).
\]

The CASH problem is mathematically formulated as minimizing the average loss across all cross-validation folds. The optimal algorithm and hyperparameter settings are obtained by minimizing this loss function. Thornton et al  \cite{thornton2013auto} were among the earliest researchers to examine the CASH problem in the AUTO-WEKA system \cite{hall2009weka}, which utilized Bayesian optimization to solve the combined task of choosing the optimal learning algorithm and adjusting hyperparameters. Bayesian optimization \cite{brochu2010tutorial} employs a probabilistic model to estimate performance of different hyperparameter configurations, thereby achieving a balance between exploring new configurations and exploiting known optimal ones. Thornton et al. \cite{thornton2013auto} investigated Bayesian tree-based search algorithms, demonstrating that the SMAC framework utilizing random forests \cite{hutter2011sequential} performs better than its predecessor \cite{bergstra2011algorithms}. Within this study, SMAC is leveraged to optimize the CASH process. Besides leveraging random forests \cite{breiman2001random}, SMAC is distinguished by its ability to Enhance cross-validation efficiency through independent fold evaluation while filtering out ineffective hyperparameter configurations early on.

Class imbalance is a common issue in classification tasks, especially in medical datasets where minority class labels often lack sufficient samples. To better identify minority classes, automated machine learning provides functionality for handling imbalanced classes. The main idea is to assign different weights to each class so that the model places greater emphasis on minority class samples during training, thereby improving overall classifier performance.

\begin{figure*}[h!]
	\centering
	\includegraphics[width=\textwidth]{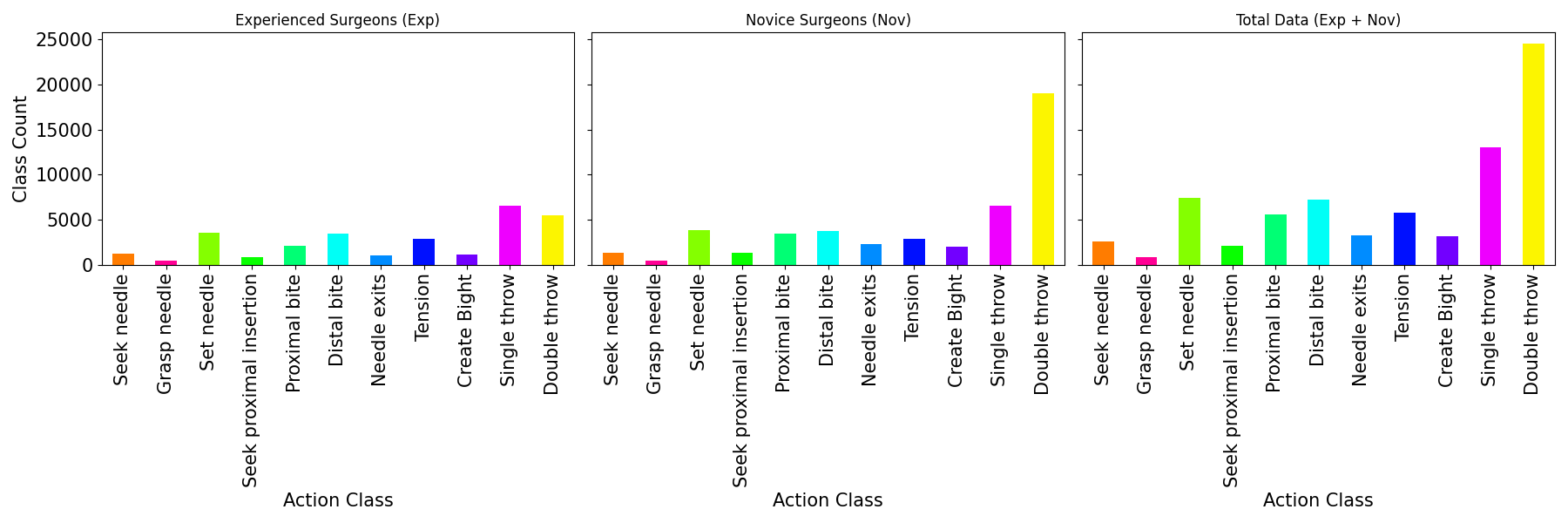}
	\caption{Class distribution across action categories shows a noticeable imbalance, with certain categories, such as Double throw and Set needle, containing significantly more samples compared to others like Grasp needle and Needle exits. This imbalance may lead to biased model performance, favoring well-represented classes while potentially underperforming on underrepresented ones.
	}
	\label{classdistribution}
\end{figure*}

In practice, AutoML combines two main approaches for handling class imbalance: \textit{sample resampling} and \textit{weighted classification}. Sample resampling includes oversampling (such as using SMOTE \cite{chawla2002smote} to produce additional data samples for minority classes) and undersampling (reducing the number of majority class samples). Weighted classification assigns higher weights to minority classes, increasing their penalty in the loss function and enhancing the model's sensitivity to these classes.

Suppose a dataset with \( C \) classes, where each class \( i \) has \( N_i \) samples, the weight for class \( i \), \( w_i \), is defined as:
\begin{equation}
	w_i = \frac{N}{C \cdot N_i}
\end{equation}
Here, \( N \) indicated total dataset cardinality. The weighted cross-entropy loss function is given by:
\begin{equation}
	L = - \sum_{i=1}^{C} w_i \cdot y_i \log(\hat{y}_i)
\end{equation}
where \( y_i \) represents the true label's one-hot encoding, and \( \hat{y}_i \) represents the model's predicted probability.

Auto-Sklearn integrates meta-learning \cite{hutter2011sequential} and automated ensemble \cite{guyon2010model, lacoste2014agnostic} techniques to accelerate model optimization and enhance overall performance. First, meta-learning leverages prior knowledge from previous tasks to predict configurations has a high probability of delivering reliable results on new data. It computes meta-features (such as dataset size, dimensionality, and class distribution) for several datasets to infer suitable algorithms for the new dataset. This process involves an offline phase and an online phase: in the offline phase, the system analyzes multiple datasets to identify high-performing configurations and stores them; when encountering a new dataset, it quickly identifies the most similar stored configuration based on meta-features as the starting point for Bayesian optimization, significantly reducing optimization time. Here we follow the Auto-Sklearn approach for our task.

Additionally, the automated ensemble method further improves robustness and performance during model optimization. Unlike traditional Bayesian optimization that seeks a single best model configuration, Auto-Sklearn retains multiple models trained during the search process and constructs an ensemble model from them. This approach avoids early convergence on a specific configuration, enhancing model robustness. The ensemble is constructed using a post-processing method that combines predictions from candidate models through weighted averaging, thereby optimizing overall performance. This method eliminates the need to fine-tune a single hyperparameter setting excessively, instead leveraging diverse model outputs to significantly boost generalization capabilities. The overall workflow for automated machine learning is as shown in Fig. \ref{framework}

\begin{figure}[ht]
	\centering
	\includegraphics[width=0.5\textwidth]{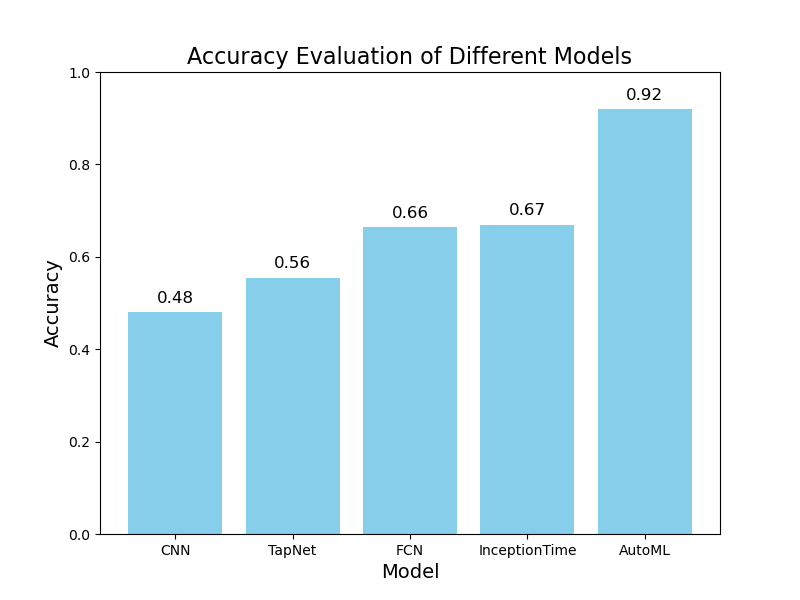}
	\caption{Model accuracy comparison, illustrating the impact of automated optimization versus conventional deep learning approaches.}
	\label{modelcomp}
\end{figure}

\begin{table*}[h!]
	\centering
	\caption{Trajectory Data Characteristics for Novice and Experienced Surgeons}
	\label{trajectorydatacharacteristics}
	\begin{tabular}{lccccc}
		\hline
		\textbf{Operation Type} & \textbf{Trajectory Length (data points)} & \textbf{Speed} & \textbf{Acceleration} & \textbf{Trajectory Range} & \textbf{Trajectory Deviation} \\
		\hline
		Novice Surgeon & 11716 & 14.59 & 113.84 & 4540.26 & 42.91 \\
		Experienced Surgeon & 4782 & 9.90 & 38.05 & 669.19 & 44.99 \\
		\hline
	\end{tabular}
\end{table*}

\begin{table*}[h]
	\centering
	\caption{Classifier Contribution Table}
	\label{tab:adjusted_classifier_contribution}
	\begin{tabular}{c|l|c|c|c|c}
		\hline
		\textbf{No.} & \textbf{Classifier} & \textbf{Ensemble Weight} & \textbf{Cost} & \textbf{Balancing Strategy} & \textbf{Validation Score} \\
		\hline
		1 & RandomForestClassifier (model 2) & 0.3 & 0.0737 & Weighting & 0.927 \\
		2 & RandomForestClassifier (model 3) & 0.06 & 0.8319 & Weighting & 0.85 \\
		3 & HistGradientBoostingClassifier (model 4) & 0.04 & 0.4264 & Weighting & 0.82 \\
		4 & KNeighborsClassifier (model 8) & 0.22 & 0.0997 & None & 0.9 \\
		5 & HistGradientBoostingClassifier (model 9) & 0.04 & 0.1113 & Weighting & 0.84 \\
		6 & RandomForestClassifier (model 13) & 0.34 & 0.0722 & Weighting & 0.93 \\
		\hline
	\end{tabular}
\end{table*}

\section{Experiments}

\begin{figure*}[h!]
	\centering
	\includegraphics[width=0.8\textwidth]{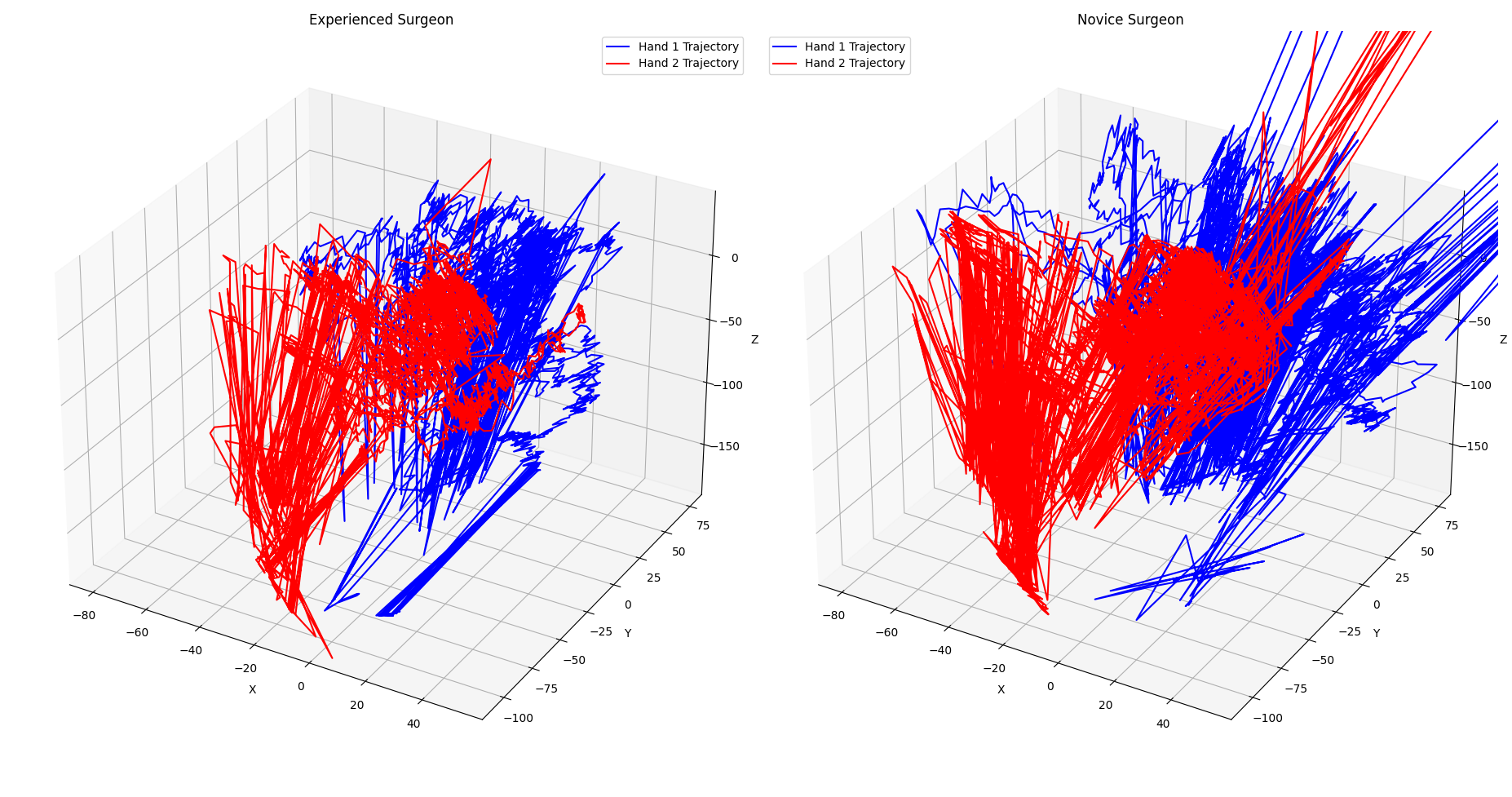}
	\caption{Experienced surgeons' trajectories show consistent, controlled, and concentrated movements within a defined spatial range, reflecting precision and coordination. In contrast, novice surgeons' trajectories display irregular and dispersed patterns, suggesting a lack of control and coordination. The trajectories of novice surgeons also demonstrate frequent abrupt deviations and scattered points, indicating variability and more erratic hand movements. These inconsistencies and outliers in novice trajectories may introduce noise to the data, potentially impacting model training by skewing predictions toward these deviations.}
	\label{trajvisual}
\end{figure*}

In this section, we utilize the laparoscopy surgical training data that we have collected from six professional and four novice laparoscopy surgeons using a laparoscopy training box. In average, the professional surgeons had 15
years’ expertise in general surgical practice and 133
hours in laparoscopic operations, and the trainee surgeons had 5 years’ surgical expertise and had gained 3 hours laparoscopic training. We recorded in total 10 videos of suturing exercise across all professional and novice participants. The experiment protocol for collection of this data was approved by the Ethics Committee of the Heriot-Watt University. The data is divided into two categories: one set was collected from operations performed by experienced surgeons, labeled as \texttt{Exp} files (6 files), while the other set was gathered from novice surgeons, labeled as \texttt{Nov} files (4 files). 
Trajectories were drawn from videos of the surgical tasks by calibrating the camera and tracking coloured markers, precisely positioned on each instrument shaft close to the instrument tip. Measurement of properties of the projected image of these markers allowed the 3D position of the markers and thus the instrument tip to be tracked across frames of the video. 
The videos had a frame rate of 25 fps and a resolution of 1920 by 1080 pixels.   
Comparison of with videos taken from a robotically controlled instrument as ground truth allowed an estimate of positioning accuracy of 5mm to be made.
These data not only capture the differences in precision and consistency but also reflect typical movement characteristics associated with different skill levels, providing a rich source of data for training and validating the surgical action recognition model. 
Our dataset comprises 11 categories of movement, as listed in Table \ref{tab:classification_report_novice}, each representing a specific surgical action. The 11 categories were created by combining surgical phase modelling \cite{lalys2014surgical} and heirarchical task analysis \cite{sarker2008constructing} to the suture task.
These activities are broadly composed of three phases:
Bringing the needle under control of the instrument and setting orientation (Seek needle, Grasp needle, Set Needle).
Passing the suture material through the tissue phantom (Seek proximal insertion, Proximal bite, Distal bite, Needle exits, Tension.
Forming a surgeon's knot (Create bight, Single Throw, Double throw).
This resulted in 11 distinct activities that must be performed in sequence to complete the task and as a result were conserved across the dataset. To fully illustrate the attributes of trajectory data, we offer a detailed computational table \ref{trajectorydatacharacteristics} of trajectory characteristics, where each attribute is calculated by averaging. Here, we don’t need to use dimensions for these attributes, because they are the same for both novices and experienced surgeons, making them easier to compare. Additionally, Fig. \ref{classdistribution} illustrates the data distribution across categories, clearly reflecting the class imbalance issue among different actions. Finally, a visualization of the trajectories in Fig. \ref{trajvisual} shows the spatial characteristics of the operations, displaying the paths taken by novice and experienced surgeons on the same task. These tables and figures help us gain an in-depth insight of the dataset, supporting subsequent model training and performance evaluation.


\begin{table}[h]
	\centering
	\caption{Classification Report on Test Dataset}
	\label{tab:classification_report_novice}
	\begin{tabular}{lcccc}
		\hline
		\textbf{Class} & \textbf{Precision} & \textbf{Recall} & \textbf{F1-score} & \textbf{Support} \\
		\hline
		Create Bight             & 0.90 & 0.90 & 0.90 & 1005 \\
		Distal bite              & 0.84 & 0.93 & 0.88 & 1861 \\
		Double throw             & 0.96 & 0.92 & 0.94 & 9604 \\
		Grasp needle             & 0.96 & 0.98 & 0.97 & 251  \\
		Needle exits             & 0.90 & 0.85 & 0.88 & 1076 \\
		Proximal bite            & 0.88 & 0.92 & 0.90 & 1754 \\
		Seek needle              & 0.99 & 0.96 & 0.97 & 684  \\
		Seek proximal insertion  & 0.85 & 0.86 & 0.85 & 630  \\
		Set needle               & 0.94 & 0.97 & 0.95 & 1876 \\
		Single throw             & 0.89 & 0.89 & 0.89 & 3259 \\
		Tension                  & 0.86 & 0.91 & 0.88 & 1432 \\
		\hline
		\textbf{Accuracy}        &       &       & 0.92 & 23432 \\
		\textbf{Macro avg}       & 0.91 & 0.92 & 0.91 & 23432 \\
		\textbf{Weighted avg}    & 0.92 & 0.92 & 0.92 & 23432 \\
		\hline
	\end{tabular}
\end{table}

\begin{figure}[h]
	\centering
	\includegraphics[width=0.5\textwidth]{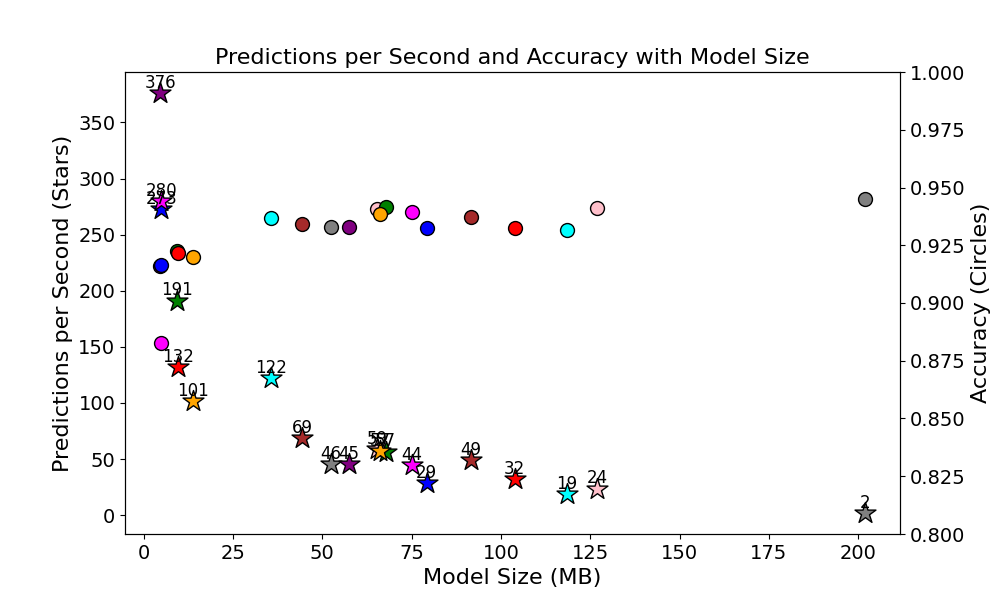}
	\caption{Analysis of prediction speed and accuracy relative to model size in ensemble architectures}
	\label{predictionvsmodelsize}
\end{figure}

The training platform used is the HP ZBook Fury 16 G10, featuring an Intel Core i9-13950HX processor, 64GB of DDR5 RAM, and an NVIDIA RTX 5000 Ada GPU with 16GB of dedicated memory. We use Auto-Sklearn \cite{feurer2015efficient} as the implementation. For data partitioning, considering the significant differences in data characteristics between the \texttt{Exp} and \texttt{Nov} datasets, we used all \texttt{Exp} data along with half of the \texttt{Nov} data as the training and validation set. The remaining half of the \texttt{Nov} data was designated as the test set. This devision ensures that the model is exposed to a diverse dataset while reserving part of the novice data for unbiased testing, highlighting the model’s capacity for adaptation across experience levels. 

\section{Results}

Fig. \ref{modelcomp} presents a comparison of various models’ effectiveness on the given dataset. AutoML achieves the highest accuracy at \textit{92\%}, significantly outperforming the classic deep learning based models. InceptionTime \cite{ismail2020inceptiontime} and FCN \cite{wang2017time} achieve comparable accuracies of \textit{67\%} and \textit{66\%}, respectively, demonstrating relatively strong performance. TapNet \cite{zhang2020tapnet} achieves an accuracy of \textit{56\%}, while CNN \cite{zhao2017convolutional} performs the worst with an accuracy of only \textit{48\%}. The superior performance of AutoML is likely due to its automated feature selection, hyperparameter optimization, and model ensembling techniques, whereas deep learning models like InceptionTime and FCN benefit from their specific adaptability to time-series data. CNN models, typically more suitable for image data, show limited modeling capacity for time-series tasks, which may explain their lowest accuracy. It is important to note that we used open-source implementations of deep learning models without any optimization tailored to our dataset. 


The overall accuracy on the novice dataset is \textit{91.7\%}. Combined with the detailed metrics in the classification report (see Table \ref{tab:classification_report_novice}), we can gain insights into the model’s performance, stability, and reliability under conditions of class imbalance.

First, there is a significant disparity in the number of samples per class, with \texttt{Double throw} having a support count of 9604, while \texttt{Grasp needle} only has 251. Such imbalance usually leads to better performance on classes with larger sample sizes, potentially at the expense of smaller classes. By examining the \textit{Weighted avg} and \textit{Macro avg} metrics, we can better understand comprehensive effectiveness of this approach. In this dataset, the weighted average F1-score is \textit{0.92}, indicating that the model maintains stable performance across most classes, especially in those with larger sample sizes, which enhances overall performance on imbalanced data. Meanwhile, the macro average F1-score is \textit{0.91}, demonstrating that the model performs relatively consistently across classes without a strong bias towards larger classes, showing a degree of generalization.

The \textit{precision} and \textit{recall} values in the classification report further highlight the model's stability and reliability. For instance, the F1-score for \textit{Double throw} is \textit{0.94}, while \textit{Grasp needle} achieves an F1-score of \textit{0.97}. This suggests that the model achieves high precision and recall across both large and small classes, reliably identifying various surgical actions. Moreover, the high macro and weighted averages confirm the model's stable performance despite class imbalance.

Finally, \textit{support} provides the sample count for each class, clearly showing the distribution differences across classes. Despite the imbalance, the model still achieves a high overall F1-score and accuracy, demonstrating its robustness and adaptability. These metrics collectively validate the model's robust performance on the novice dataset, providing reliable support for surgical action recognition applications.

As shown in Table \ref{tab:adjusted_classifier_contribution}, notable variations exist in the contribution and performance of different classifiers within the ensemble model. In terms of interpretability, models like RandomForestClassifier and KNeighborsClassifier generally provide higher interpretability, which aids in analyzing the specific role of each model in the predictions. The table's Ensemble Weight attribute indicates contribution of each model to the ensemble; for example, RandomForestClassifier (model 13) has the highest weight (0.34), showing that it plays a dominant role in the final predictions. Additionally, the Balancing Strategy column shows the balancing strategy for each model, with most models using the Weighting strategy to address data imbalance issues, enhancing robustness and fairness in predictions. Finally, the Validation Score column reflects the models' performance on the validation set, where models with higher weights tend to have higher validation scores, further emphasizing their importance in the overall ensemble. Relationship between model size, prediction speed and prediction accuracy is as shown in Fig. \ref{predictionvsmodelsize}. This figure illustrates how the size of ensemble models (in MB) affects predictions per second and prediction accuracy. Smaller models exhibit significantly higher prediction speeds, with speeds exceeding 100 predictions per second for models under 25MB, making them highly suitable for scenarios with strict real-time requirements. However, as the model size increases, prediction speed drops significantly, while accuracy shows a gradual upward trend. Notably, the largest model (200MB) achieves a high accuracy close to 0.95 but suffers from a substantial decline in prediction speed, making it less viable for real-time applications. A closer look reveals that models within the 10MB to 25MB range strike the best balance between speed and accuracy. These models deliver prediction speeds exceeding 100 predictions per second while maintaining strong accuracy, making them ideal for scenarios that demand both real-time performance and reliable predictions. This range highlights a practical trade-off, ensuring sufficient speed without compromising predictive performance, making them highly suitable for scenarios with strict real-time requirements.

\section{CONCLUSIONS and Future Works}

In this study, we developed a traditional machine learning based and fast deployable laparoscopic surgical action recognition system, effectively addressing challenges related to data imbalance, model interpretability, stability, and reliability. By training our model on a high-performance platform and leveraging the automated machine learning framework, we achieved high accuracy and stable performance across multiple surgical action categories. Additionally, the classifier contribution table and model performance analysis provided valuable insights into model selection and optimization, validating the potential of our approach to enhance surgical action recognition and training efficiency. For upcoming studies, we plan to expand our dataset by collecting more surgical action data to improve model robustness. We also intend to explore the latest deep learning based detection techniques to further enhance the system's accuracy and real-time capabilities.

\addtolength{\textheight}{-12cm}   



%

\section*{Acknowledgment}

This research has been funded by the Engineering and Physical Sciences Research Council (EPSRC) of United Kingdom under Grant Reference EP/Y017307/1.

\bibliographystyle{IEEEtran}
\bibliography{IEEEabrv,ref}

\end{document}